# *Research Note*
# A Complete Classification of Tractability in RCC-5

**Peter Jonsson**                                    PETEJ@IDA.LIU.SE
**Thomas Drakengren**                                THODR@IDA.LIU.SE
*Department of Computer and Information Science, Linköping University*
*S-581 83 Linköping, Sweden*

## Abstract

We investigate the computational properties of the spatial algebra RCC-5 which is a restricted version of the RCC framework for spatial reasoning. The satisfiability problem for RCC-5 is known to be NP-complete but not much is known about its approximately four billion subclasses. We provide a complete classification of satisfiability for all these subclasses into polynomial and NP-complete respectively. In the process, we identify all maximal tractable subalgebras which are four in total.

## 1. Introduction

Qualitative spatial reasoning has received a constantly increasing amount of interest in the literature. The main reason for this is, probably, that spatial reasoning has proved to be applicable to real-world problems in, for example, geographical database systems (Egenhofer, 1991; Grigni, Papadias, & Papadimitriou, 1995) and molecular biology (Cui, 1994). In both these applications, the size of the problem instances can be huge, so the complexity of problems and algorithms is a highly relevant area to study. However, questions of computational complexity have not received so much attention in the literature; two notable exceptions are the results reported by Nebel (1995) and Renz and Nebel (1997). In this article we take a small step towards a better understanding of complexity issues in qualitative spatial reasoning.

A well-known framework for qualitative spatial reasoning is the so-called RCC approach (Randell & Cohn, 1989; Randell, Cui, & Cohn, 1992). This approach is based on modelling qualitative spatial relations between regions using first-order logic. Of special interest, from a complexity-theoretic standpoint, are the two subclasses RCC-5 and RCC-8. It is well-known that both RCC-5 and RCC-8 have quite weak expressive power. Although they can be used to describe spatial situations, they are very general and should perhaps better be described as topological algebras. However, we will denote these algebras as spatial algebras in order to avoid terminological confusion; the term topological algebra has a well-established but completely different meaning in mathematics (Mallios, 1986).

Bennett (1994) has shown the sufficiency of using propositional logics for reasoning about RCC-5 and RCC-8. Hence, the reasoning becomes more efficient when compared to reasoning in a full first-order logic. Bennett's approach uses classical propositional logic for RCC-5 and intuitionistic propositional logic for RCC-8. Unfortunately, these logics are known to be computationally hard. The satisfiability problem for classical propositional logic and intuitionistic propositional logic is NP-complete (Cook, 1971) and PSPACE-complete (Statman, 1979) respectively. However, the complexity of the underlying logic does not carry over in both cases; Renz and Nebel (1997) have shown that the satisfiability problem for both RCC-5 and RCC-8 is NP-complete. The full proofs can be found in (Renz, 1996).

These findings motivate the search for tractable subclasses of RCC-5 and RCC-8. Nebel (1995) showed that reasoning with the basic relations in RCC-8 is a polynomial-time problem. Renz and Nebel (1997) improved this result substantially by showing the following results:

- There exists a large, maximal subclass of RCC-8, denoted $\widehat{\mathcal{H}}_8$, which contains all basic relations and is polynomial. $\widehat{\mathcal{H}}_8$ contains 148 elements out of 256 (58%).

- There exists a large, maximal subclass of RCC-5, denoted $\widehat{\mathcal{H}}_5$, which contains all basic relations and is polynomial. $\widehat{\mathcal{H}}_5$ contains 28 elements out of 32 (87%). Furthermore, this is the unique, maximal subclass of RCC-5 containing all basic relations.

We will concentrate on RCC-5 in this article. The main result is a complete classification of all subclasses of RCC-5 with respect to tractability. The classification makes it possible to determine whether a given subclass is tractable or not by a simple test that can be carried out by hand or automatically. We have thus gained a clear picture of the tractability borderline in RCC-5. As is more or less necessary when showing results of this kind, the main proof relies on a case analysis performed by a computer. The number of cases considered was roughly $4 \times 10^4$. The analysis cannot, of course, be reproduced in a research paper or be verified manually. Hence, we include a description of the programs used. The programs are also available as an on-line appendix to this article.

The structure of the article is as follows: Section 2 defines RCC-5 and some auxiliary concepts. Section 3 contains the tractability proofs for three subclasses of RCC-5. In Section 4 we show that these subclasses together with $\widehat{\mathcal{H}}_5$ are the only maximal tractable subclasses of RCC-5. The article concludes with a brief discussion of the results.

## 2. The RCC-5 Algebra

We follow Bennett (1994) in our definition of RCC-5. RCC-5 is based on the notions of *regions* and *binary relations* on them. A region $p$ is a variable interpreted over the non-empty subsets of some fixed set. It should be noted that we do not require the sets to be open sets in some topological space. This is no limitation since it is impossible to distinguish interior points from boundary points in RCC-5. Thus we can take any set $\mathcal{X}$ and use the discrete topology $\mathcal{T} = \langle \mathcal{X}, 2^{\mathcal{X}} \rangle$, where every subset of $\mathcal{X}$ is an open set in $\mathcal{T}$.

We assume that we have a fixed universe of variable names for regions. Then, an *R-interpretation* is a function that maps region variables to the non-empty subsets of some set.

Given two interpreted regions, their relation can be described by exactly one of the elements of the set **B** of five *basic RCC-5 relations*. The definition of these relations can be found in Table 1. Figure 1 shows 2-dimensional examples of the relations in RCC-5. A formula of the form $XBY$ where $X$ and $Y$ are regions and $B \in \mathbf{B}$, is said to be satisfied by an $R$-interpretation iff the interpretation of the regions satisfies the relations specified in Table 1.

To express indefinite information, unions of the basic relations are used, written as sets of basic relations, leading to $2^5$ binary RCC-5 relations. Naturally, a set of basic relations is to be interpreted as a disjunction of the basic relations. The set of all RCC-5 relations $2^{\mathbf{B}}$ is denoted by $R_5$. Relations of special interest are the *null* relation $\varnothing$ (also denoted by $\bot$) and the *universal* relation **B** (also denoted $\top$).

A formula of the form $X\{B_1, \ldots, B_n\}Y$ is called an RCC-5 formula. Such a formula is satisfied by an $R$-interpretation $\Im$ iff $XB_iY$ is satisfied by $\Im$ for some $i$, $1 \leq i \leq n$. A finite set $\Theta$ of RCC-5 formulae is said to be *R-satisfiable* iff there exists an $R$-interpretation $\Im$ that satisfies every formula of $\Theta$. Such a satisfying $R$-interpretation is called an *R-model* of $\Theta$. Given an $R$-interpretation $\Im$ and a variable $v$, we write $\Im(v)$ to denote the value of $v$ under the interpretation $\Im$.

The reasoning problem we will study is the following:

INSTANCE: A finite set $\Theta$ of RCC-5 formulae.
QUESTION: Does there exist an $R$-model of $\Theta$?

$$X\{\texttt{DR}\}Y \quad \text{iff} \quad X \cap Y = \varnothing$$
$$X\{\texttt{PO}\}Y \quad \text{iff} \quad \exists a, b, c : a \in X, a \notin Y, b \in X, b \in Y, c \notin X, c \in Y$$
$$X\{\texttt{PP}\}Y \quad \text{iff} \quad X \subset Y$$
$$X\{\texttt{PPI}\}Y \quad \text{iff} \quad X \supset Y$$
$$X\{\texttt{EQ}\}Y \quad \text{iff} \quad X = Y$$

Table 1: The five basic relations of RCC-5.

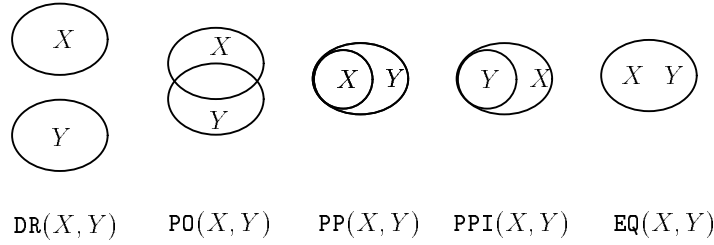

$\quad\texttt{DR}(X,Y) \quad\quad \texttt{PO}(X,Y) \quad\quad \texttt{PP}(X,Y) \quad\quad \texttt{PPI}(X,Y) \quad\quad \texttt{EQ}(X,Y)$

Figure 1: Pictorial example of the relations in RCC-5.

We denote this problem by RSAT. In the following, we often consider restricted versions of RSAT where the relations used in formulae in $\Theta$ are only from a subset $S$ of $R_5$. In this case we say that $\Theta$ is a set of formulae over $S$ and we use a parameter in the problem description to denote the subclass under consideration, $e.g.$, RSAT($S$). Note that an RSAT problem instance can be represented by a labelled directed graph, where the nodes are region variables and the arcs are labelled by relations between variables. Given an instance $\Theta$ of RSAT, we say that such a graph is a *graph representation* of $\Theta$.

We continue by defining an algebra over the RCC-5 relations.

**Definition 2.1** Let $\mathbf{B} = \{\texttt{DR}, \texttt{PO}, \texttt{PP}, \texttt{PPI}, \texttt{EQ}\}$. The RCC-5 algebra consists of the set $R_5 = 2^{\mathbf{B}}$ and the operations unary *converse* (denoted by $\smile$), binary *intersection* (denoted by $\cap$) and binary *composition* (denoted by $\circ$). They are defined as follows:

$$\forall X, Y: \quad XR^{\smile}Y \quad \text{iff} \quad YRX$$
$$\forall X, Y: \quad X(R \cap S)Y \quad \text{iff} \quad XRY \wedge XSY$$
$$\forall X, Y: \quad X(R \circ S)Y \quad \text{iff} \quad \exists Z : (XRZ \wedge ZSY)$$

If $\mathcal{S}$ is a subset of $R_5$, $\mathcal{S}$ is said to be a *subalgebra* of RCC-5 iff $\mathcal{S}$ is closed under converse, intersection and composition. It can easily be verified that $R \circ S = \bigcup \{B \circ B' | B \in R, B' \in S\}$, $i.e.$, composition is the union of the component-wise composition of basic relations.

Next, we introduce a *closure* operation. The closure operation transforms a given subclass of $R_5$ to one that is polynomially equivalent to the original subclass with respect to satisfiability. The operation is similar to the closure operation for RCC-5 introduced by Renz (1996) but it does not pose the same restrictions on the given subclass. (Renz's operation requires $\{\texttt{EQ}\}$ to be a member of the subclass to be closed.)

**Definition 2.2** Let $\mathcal{S} \subseteq R_5$. Then we denote by $\overline{\mathcal{S}}$ the *closure* of $\mathcal{S}$, defined as the least subalgebra containing $\mathcal{S}$ closed under converse, intersection and composition.

Observe that a subset $\mathcal{S}$ of $R_5$ is a subalgebra iff $\mathcal{S} = \overline{\mathcal{S}}$.

The next lemma is given without proof. A proof of the analogous result for Allen's algebra can be found in Nebel and Bürckert (1995).

**Lemma 2.3** Let $\mathcal{S} \subseteq R_5$. Then $\mathrm{RSAT}(\overline{\mathcal{S}})$ can be polynomially transformed to $\mathrm{RSAT}(\mathcal{S})$ and vice versa.

**Corollary 2.4** Let $\mathcal{S} \subseteq R_5$. $\mathrm{RSAT}(\mathcal{S})$ is polynomial iff $\mathrm{RSAT}(\overline{\mathcal{S}})$ is polynomial. $\mathrm{RSAT}(\mathcal{S})$ is NP-complete iff $\mathrm{RSAT}(\overline{\mathcal{S}})$ is NP-complete.

## 3. Tractable Subclasses of RCC-5

We begin this section by defining four tractable subalgebras of RCC-5, which can be found in Table 2. Later on, we show that these algebras are the only maximal tractable subalgebras of RCC-5. The tractability of the first algebra, $R_5^{28}$, has been established by Renz and Nebel (1997). The name $R_5^{28}$ reflects the fact that the algebra contains 28 elements.

**Theorem 3.1** $\mathrm{RSAT}(R_5^{28})$ is polynomial.

The tractability of our second algebra, $R_5^{20}$, can be settled quite easily. The algorithm can be found in Figure 2.

**Lemma 3.2** Let $\Theta$ be an instance of $\mathrm{RSAT}(R_5^{20})$. The algorithm $A^{20}$ accepts on input $\Theta$ iff $\Theta$ has an $R$-model.

**Proof:** *if:* We show the contrapositive, *i.e.*, if $A^{20}$ rejects then $\Theta$ has no $R$-model. Clearly, the satisfiability of $\Theta$ is preserved under the transformations made in lines 7-10. Note that if $XRX \in \Theta$ then $\mathtt{EQ} \in R$ if $\Theta$ is satisfiable. Thus $\Theta$ is not satisfiable if the algorithm rejects in line 5. Similarly, $\Theta$ is not satisfiable if the algorithm rejects in line 6.

*only-if:* Consider the set $\Theta$ after the completion of line 11. We denote this set by $\Theta'$. Obviously, $\Theta'$ is satisfiable if the initial $\Theta$ was satisfiable. Also observe that line 7 ensures that $\Theta'$ does not relate any variables with $\mathtt{EQ}$. Furthermore, line 8 guarantees that there is at most one relation that relates two variables.

Now, we construct an $R$-model $M$ for $\Theta'$ as follows: Let $V$ be the set of variables in $\Theta'$. Let $M$ assign non-empty sets that are pairwise disjoint to the members of $V$. Let $\mathcal{U} = \bigcup_{X \in V} M(X)$. Introduce a set of values $U' = \{\alpha_{X,Y} \mid X, Y \in V\}$ satisfying the following:

1. $\alpha_{X,Y} = \alpha_{Z,W}$ iff $X = Z$ and $Y = W$; and

2. for arbitrary $X, Y \in V$, $\alpha_{X,Y} \notin \mathcal{U}$.

For each relation of the type $X\{\mathtt{PO}\}Y$ or $X\{\mathtt{PO}, \mathtt{EQ}\}Y$, extend the sets $M(X)$ and $M(Y)$ with the element $\alpha_{X,Y}$.

Clearly, two sets $X, Y$ are disjoint (and are thus related by $\mathtt{DR}$) under $M$ unless $X\{\mathtt{PO}\}Y$ or $X\{\mathtt{PO}, \mathtt{EQ}\}Y$ is in $\Theta$. But in these cases, $X$ and $Y$ must not be disjoint. In fact, by introducing $\alpha_{X,Y}$, we have forced $X\{\mathtt{PO}\}Y$ to hold under $M$ which satisfies formulae of the type $X\{\mathtt{PO}\}Y$ as well as formulae of the type $X\{\mathtt{PO}, \mathtt{EQ}\}Y$. Hence, $M$ is an $R$-model of $\Theta'$ which implies the $R$-satisfiability of $\Theta$. □

**Theorem 3.3** $\mathrm{RSAT}(R_5^{20})$ is polynomial.

**Proof:** Algorithm $A^{20}$ correctly solves the $\mathrm{RSAT}(R_5^{20})$ problem by the previous lemma. Furthermore, the number of iterations is bounded from above by the number of variables and the number of formulae in the given instance and the tests can easily be performed in polynomial time. □

Next we show the tractability of $\mathrm{RSAT}(R_5^{17})$.

|  | $R_5^{28}$ | $R_5^{20}$ | $R_5^{17}$ | $R_5^{14}$ |
|---|---|---|---|---|
| $\perp$ | • | • | • | • |
| {DR} | • | • |  |  |
| {PO} | • | • |  |  |
| {DR, PO} | • | • |  |  |
| {PP} | • |  |  | • |
| {DR, PP} | • | • |  |  |
| {PO, PP} | • |  |  |  |
| {DR, PO, PP} | • | • |  |  |
| {PPI} | • |  |  | • |
| {DR, PPI} | • | • |  |  |
| {PO, PPI} | • |  |  |  |
| {DR, PO, PPI} | • | • |  |  |
| {PP, PPI} |  |  |  | • |
| {DR, PP, PPI} |  | • |  | • |
| {PO, PP, PPI} | • |  |  | • |
| {DR, PO, PP, PPI} | • | • |  | • |
| {EQ} | • | • | • | • |
| {DR, EQ} | • | • | • |  |
| {PO, EQ} | • | • | • |  |
| {DR, PO, EQ} | • | • | • |  |
| {PP, EQ} | • |  | • | • |
| {DR, PP, EQ} | • | • | • |  |
| {PO, PP, EQ} | • |  | • |  |
| {DR, PO, PP, EQ} | • | • | • |  |
| {PPI, EQ} | • |  | • | • |
| {DR, PPI, EQ} | • | • | • |  |
| {PO, PPI, EQ} | • |  | • |  |
| {DR, PO, PPI, EQ} | • | • | • |  |
| {PP, PPI, EQ} |  |  | • | • |
| {DR, PP, PPI, EQ} |  | • | • | • |
| {PO, PP, PPI, EQ} | • |  | • | • |
| $\top$ | • | • | • | • |

Table 2: The maximal tractable subalgebras of RCC-5.

**Theorem 3.4** RSAT($R_5^{17}$) is polynomial.

**Proof:** Consider the algorithm $A^{17}$ in Figure 2. If there exist $X, Y$ such that $X \perp Y \in \Theta$ then $\Theta$ is not satisfiable. Otherwise, we can let all variables have the same value. Since EQ is a member of every relation that occurs in $\Theta$, this interpretation is an $R$-model of $\Theta$. $\square$

We continue by proving that RSAT($R_5^{14}$) is a tractable problem. Let

$$R_5^9 = \{\{\text{PP}, \text{EQ}\}\} \cup \{R \cup \{\text{PP}, \text{PPI}\} \mid R \in R_5\}.$$

Using a machine-assisted proof, it can be shown that $R_5^{14} = \overline{R_5^9}$ so it is sufficient to prove the tractability of RSAT($R_5^9$) by Corollary 2.4. The program that we used for showing this is available as an on-line appendix to this article.

From now on, let $\Theta$ be an arbitrary instance of RSAT($R_5^9$) and $G = \langle V, E \rangle$ be its graph representation. The following proofs are similar in spirit to some of the proofs appearing in Drakengren and

```
1  algorithm A^20
2  Input: An instance Θ of RSAT(R_5^20).
3   repeat
4    Θ' ← Θ
5    if ∃X, R : XRX ∈ Θ and EQ ∉ R  then reject
6    if ∃X, Y : X⊥Y ∈ Θ  then reject
7    if ∃X, Y : X ≠ Y and X{EQ}Y ∈ Θ  then substitute Y for X in Θ
8    if ∃X, Y, R, S : XRY ∈ Θ and XSY ∈ Θ  then
9       Θ ← (Θ − {XRY, XSY}) ∪ {X(R ∩ S)Y}
10   if ∃X, R : XRX ∈ Θ and EQ ∈ R  then Θ ← Θ − {XRX}
11  until Θ = Θ'
12  accept

1  algorithm A^17
2  Input: An instance Θ of RSAT(R_5^17).
3  if ∃X, Y such that X⊥Y ∈ Θ  then reject
4  else accept

1  algorithm A^9
2  Input: An instance Θ of RSAT(R_5^9) with graph representation G.
3  Let G' be the graph obtained from G by removing arcs which are not labelled {PP, EQ}.
4  Find all strongly connected components C in G'
5  for every arc e in G whose relation does not contain EQ  do
6     if e connects two nodes in some C  then reject
7  accept
```

Figure 2: Algorithms for $\text{RSAT}(R_5^{20})$, $\text{RSAT}(R_5^{17})$ and $\text{RSAT}(R_5^9)$.

Jonsson (1996). The algorithm itself is reminiscent of an algorithm by van Beek (1992) for deciding satisfiability in the point algebra.

**Definition 3.5** A RCC-5 relation $R$ is said to be an *acyclic* relation iff any cycle in any $G$ with $R$ on every arc is never satisfiable.

The relation PP is an example of an acyclic relation while {PP, EQ} is not acyclic. We continue by showing a few properties of acyclic relations.

**Proposition 3.6** Let $R$ be an acyclic relation. Then every relation $R' \subseteq R$ is acyclic.

**Proof:** Since taking subsets of $R$ constrains satisfiability further, the result follows. □

**Proposition 3.7** Let $R$ be an acyclic relation, and choose $A$ such that $A \subseteq \{R' \mid R' \subseteq R\}$. Then, any cycle in $G$ where every arc is labelled by some relation in $A$ is unsatisfiable.

**Proof:** Same argument as in the previous proposition. □

The following definition is needed in the following proofs.

**Definition 3.8** Let $\mathcal{I}$ be an instance of the $R$-satisfiability problem, $M$ a model for $\mathcal{I}$, and $r \in R_5$ a relation between two region variables $X$ and $Y$ in $\mathcal{I}$. Then $r$ is said to be *satisfied as $r'$* in $M$ for any relation $r' \subseteq r$, such that $Xr'Y$ is satisfied in $M$.

The definition may seem a bit cumbersome but the essence should be clear. As an example, let $X$ and $Y$ be region variables related by $X\{\texttt{PO},\texttt{PP}\}Y$, and $M$ a model where $X$ is interpreted as $\{1,2\}$ and $Y$ as $\{1,2,3\}$. Then in $M$, $\{\texttt{PO},\texttt{PP}\}$ is satisfied as $\{\texttt{PP}\}$, but also as $\{\texttt{PO},\texttt{PP}\}$.

**Lemma 3.9** Let $R$ be an acyclic relation, and $A, A'$ sets such that $A \subseteq \{R' \mid R' \subseteq R\}$ and $A' \subseteq \{a \cup \{\texttt{EQ}\} \mid a \in A\}$. Then, every cycle $C$ labelled by relations in $A \cup A'$ is satisfiable iff it contains only relations from $A'$. Furthermore, all relations in the cycle have to be satisfied as $\texttt{EQ}$.

**Proof:** *only-if:* Suppose that a cycle $C$ is satisfiable and that it contains some relation from $A$. Apply induction on the number $n$ of arcs in the cycle. For $n = 1$, we get a contradiction by Proposition 3.7. So, suppose for the induction that $C$ contains $n + 1$ arcs. Let $M$ be an $R$-model for the relations in $C$. It cannot be the case that every relation in $C$ is satisfied in $M$ as some relation in $A$, by Proposition 3.7. Thus, some relation $R'$ in $C$ has to be satisfied as $\texttt{EQ}$. But then we can collapse the two variables connected by $R'$ to one variable, and we have a cycle with $n$ nodes containing a relation from $A$. This contradicts the induction hypothesis.

*if:* Suppose that a cycle $C$ contains only relations in $A'$. Then $C$ can be satisfied by choosing $\texttt{EQ}$ on every arc. Notice that the only-if part implies that $C$ *must* be satisfied by choosing $\texttt{EQ}$ on every arc. Hence, the variables are forced to be equal. $\square$

After having studied acyclic relations, we will now turn our attention to *DAG-satisfying* relations. The formal definition is as follows.

**Definition 3.10** A basic relation $B$ is said to be *DAG-satisfying* iff any DAG (directed acyclic graph) labelled only by relations containing $B$ is satisfiable, *i.e.*, if the corresponding RSAT problem has a model.

A typical example of a DAG-satisfying relation is $\texttt{EQ}$. Given a DAG labelled only by relations containing $\texttt{EQ}$, we can always satisfy these relations by assigning some non-empty set $S$ to all variables.

We can now show that $\texttt{PP}$ is a DAG-satisfying relation.

**Definition 3.11** Let $G$ be an arbitrary DAG. A node $v$ in $G$ is said to be a *terminal* node iff there are no arcs which start in $v$.

**Lemma 3.12** The basic relation $\texttt{PP}$ is DAG-satisfying.

**Proof:** Let $G$ be a DAG labelled only by relations containing $\texttt{PP}$. We show that $G$ is satisfied by some $R$-model $M$. Induction on $n$ which is the number of nodes in $G$. The case when $n = 1$ is trivial. Suppose that $G$ has $n + 1$ nodes and remove a terminal node $g$. By induction, the remaining graph $G' = \langle V', E' \rangle$ is satisfiable by a model $M'$. We shall now construct a model $M$ of $G$, which agrees with $M'$ on every variable in $G'$. Let $S = \bigcup \{M'(v) \mid v \in V'\}$ and let $\alpha$ be an element not in $S$. Let $M(g) = S \cup \{\alpha\}$. Obviously, $M$ is a model of $G$. $\square$

We now state a simple result from Drakengren and Jonsson (1996).

**Lemma 3.13** Let $G$ be irreflexive[1] and have an acyclic subgraph $D$. Then those arcs of $G$ which are not in $D$ can be reoriented so that the resulting graph is acyclic.

By specializing this result, we get the next lemma.

**Lemma 3.14** Let $G$ be irreflexive with an acyclic subgraph $D$ and let the arcs of $D$ be labelled by relations containing $\texttt{PP}$, and the arcs not in $D$ being labelled by relations containing $\texttt{PP}$ and $\texttt{PPI}$. Then $G$ is $R$-satisfiable.

---
1. A graph is irreflexive iff it has no arcs from a node $v$ to the node $v$.

**Proof:** Reorient the arcs of $G$ such that the resulting graph is acyclic. This is always possible by the previous lemma. Furthermore, whenever an arc is reoriented, also invert the relation on that arc, so that $G'$ is satisfiable iff $G$ is. By this construction, only arcs containing both PP and PPI have been reoriented, so every arc in the DAG $G'$ contains PP and, thus, since PP is DAG-satisfying by Lemma 3.12, $G'$ is satisfiable. Consequently, $G$ is also satisfiable. □

**Lemma 3.15** Algorithm $A^9$ correctly solves $\text{RSAT}(R_5^9)$.

**Proof:** Assume that the algorithm finds a strongly connected component of $G'$ (which then contains only the relation $\{\text{PP}, \text{EQ}\}$), containing two nodes that in $G$ are connected by an arc $e$ that is labelled by a relation $R'$ which does not contain EQ. Then there exists a cycle $C$ in which the relation of every arc contains EQ, such that $e$ connects two nodes in that $C$ but $e$ is not part of that cycle. By Lemma 3.9, $C$ can be satisfied only by choosing the relation EQ on every arc in $C$, and since $R'$ does not admit EQ, $C$ is unsatisfiable.

Otherwise, every such strongly connected component can be collapsed to a single node, removing all arcs which start and end in the collapsed node. This transformation retains satisfiability using the same argument as above. After collapsing, the subgraph obtained by considering only arcs labelled $\{\text{PP}, \text{EQ}\}$ will be acyclic. Since the remaining arcs are labelled by relations containing both PP and PPI, the graph is $R$-satisfiable by Lemma 3.14. (Note that the graph will be irreflexive since every node is contained in some strongly connected component.) □

**Lemma 3.16** Given a graph $G = \langle V, E \rangle$, algorithm $A^9$ runs in $O(|V| + |E|)$ time.

**Proof:** Strongly connected components can be found in $O(|V| + |E|)$ time (Baase, 1988) and the remaining test can also be made in $O(|V| + |E|)$ time. □

**Theorem 3.17** $\text{RSAT}(R_5^{14})$ can be solved in polynomial time.

**Proof:** $\text{RSAT}(R_5^9)$ is polynomial by the previous two lemmata. Since $R_5^{14} = \overline{R_5^9}$, $\text{RSAT}(R_5^{14})$ can be solved in polynomial time by Corollary 2.4. □

## 4. Classification of RCC-5

Before we can give the classification of RCC-5 we need two NP-completeness results.

**Theorem 4.1** $\text{RSAT}(\mathcal{S})$ is NP-complete if

1. (Renz & Nebel, 1997) $\mathcal{C}_1 = \{\{\text{PO}\}, \{\text{PP}, \text{PPI}\}\} \subseteq \mathcal{S}$, or

2. $\mathcal{C}_2 = \{\{\text{DR}, \text{PO}\}, \{\text{PP}, \text{PPI}\}\} \subseteq \mathcal{S}$.

**Proof:** The proof for $\mathcal{C}_2$ is by polynomial-time reduction from $\text{RSAT}(\mathcal{C}_1)$. Let $\Theta$ be an arbitrary instance of $\text{RSAT}(\mathcal{C}_1)$. Construct the following set:

$$\Theta' = \{X\{\text{PP}, \text{PPI}\}Y \mid X\{\text{PP}, \text{PPI}\}Y \in \Theta\} \cup \{X\{\text{DR}, \text{PO}\}Y \mid X\{\text{PO}\}Y \in \Theta\}.$$

Clearly, $\Theta'$ can be obtained from $\Theta$ in polynomial time and $\Theta'$ is an instance of $\text{RSAT}(\mathcal{C}_2)$. We show that $\Theta$ is satisfiable iff $\Theta'$ is satisfiable.

*only-if:* Assume that there exists an $R$-model $I$ of $\Theta$. It is not hard to see that $I$ is also an $R$-model of $I'$ since if $X\{\text{PO}\}Y$ under $I$ then $X\{\text{DR}, \text{PO}\}Y$ under $I$. Thus $\Theta'$ is $R$-satisfiable if $\Theta$ is $R$-satisfiable.

*if:* Assume the existence of an $R$-model $I'$ that assigns subsets of some set $\mathcal{U}$ to the region variables of $\Theta'$. Let $\alpha$ be an element such that $\alpha \notin \mathcal{U}$. We construct a new interpretation $I$ as follows: $I(x) = I'(x) \cup \{\alpha\}$ for every variable $x$ in $\Theta'$. It can easily be seen that the following holds for $I$:

A COMPLETE CLASSIFICATION OF TRACTABILITY IN RCC-51. If $x\{\texttt{DR}\}y$ under $I'$ then $x\{\texttt{PO}\}y$ under $I$.

2. If $x\{\texttt{PO}\}y$ under $I'$ then $x\{\texttt{PO}\}y$ under $I$.

3. If $x\{\texttt{PP}\}y$ under $I'$ then $x\{\texttt{PP}\}y$ under $I$.

4. If $x\{\texttt{PPI}\}y$ under $I'$ then $x\{\texttt{PPI}\}y$ under $I$.

It is easy to see that if $x\{\texttt{PP},\texttt{PPI}\}y$ under $I'$ then $x\{\texttt{PP},\texttt{PPI}\}y$ under $I$. Similarly, if $x\{\texttt{DR},\texttt{PO}\}y$ under $I'$ then $x\{\texttt{PO}\}y$ under $I$. It follows that $I$ is a model of $\Theta$ so $\Theta$ is $R$-satisfiable if $\Theta'$ is $R$-satisfiable. $\square$

The main theorem can now be stated and proved.

**Theorem 4.2** For $\mathcal{S} \subseteq R_5$, $\text{RSAT}(\mathcal{S})$ is polynomial iff $\mathcal{S}$ is a subset of some member of $R_P = \{R_5^{28}, R_5^{20}, R_5^{17}, R_5^{14}\}$, and NP-complete otherwise.

**Proof:** *if:* For each $R \in R_P$, $\text{RSAT}(R)$ is polynomial as was shown in the previous section.
*only-if:* Choose $\mathcal{S} \subseteq R_5$ such that $\mathcal{S}$ is not a subset of any algebra in $R_P$. For each subalgebra $R \in R_P$, choose a relation $x$ such that $x \in \mathcal{S}$ and $x \notin R$. This can always be done since $\mathcal{S} \not\subseteq R$. Let $X$ be the set of these relations and note that $X$ is not a subset of any algebra in $R_P$. The set $R_P$ contains four algebras so by the construction of $X$, $|X| \leq 4$. Observe that $\text{RSAT}(\mathcal{S})$ is NP-complete if $\text{RSAT}(X)$ is NP-complete.

To show that $\text{RSAT}(\mathcal{S})$ has to be NP-complete, a machine-assisted case analysis of the following form was performed:

1. Generate all subsets of $R_5$ of size $\leq 4$. There are $\sum_{i=0}^{4} \binom{32}{i} = 41449$ such subsets.

2. Let $\mathcal{T}$ be such a set. Test: $\mathcal{T}$ is a subset of some subalgebra in $R_P$ or $\mathcal{C}_i \subseteq \overline{\mathcal{T}}$ for some $i \in \{1,2\}$.

The test succeeds for all $\mathcal{T}$. Hence, $\text{RSAT}(\mathcal{S})$ is NP-complete by Corollary 2.4. $\square$

The program used for showing the previous theorem appears in the on-line appendix of this article.

## 5. Discussion

The main problem of reporting tractability results for restricted classes of problems is the difficulty of isolating interesting and relevant subclasses. The systematic approach of building complete classifications is a way of partially overcoming this problem. If the problem class under consideration is regarded relevant, then its tractable subclasses should be regarded relevant if the computational problem is of interest. This is especially true in spatial reasoning where the size of the problem instances can be extremely large; one good example is spatial reasoning in connection with the Human Genome project (Cui, 1994).

Another advantage with complete classifications is that they are more satisfactory from a scientific point of view; to gain a clear picture of the borderline between tractability and intractability has an intrinsic scientific value. A common critique is that complete classifications tend to generate certain classes which are totally useless. For instance, the subalgebra $R_5^{17}$ is certainly of no use. It must be made clear that such critique is unjustified since the researcher who makes a complete classification does not deliberately invent useless classes. Instead, if useless classes appear in a complete classification, they are unavoidable parts of the classification.

219

The work reported in this article can be extended in many different ways. One obvious extension is to study other computational problems than the RSAT problem. Renz (1996) has studied two problems, RMIN and RENT, on certain subclasses of RCC-5 and RCC-8. The RMIN problem is to decide if a set of spatial formulae $\Theta$ is minimal, *i.e.*, whether all basic relations in every formula of $\Theta$ can be satisfied or not. The RENT problem is to decide whether a formula $XRY$ is entailed by a set of spatial formulae. Grigni et al. (1995) study a stronger form of satisfiability which they refer to as *realizability*: A finite set $\Theta$ of RCC-5 formulae is said to be realizable iff there exist regions on the plane bounded by Jordan curves which satisfy the relations in $\Theta$. Grigni et al. (1995) have shown that the realizability problem is much harder than the satisfiability problem. For instance, deciding realizability of formulae constructed from the two relations DR and PO is NP-complete while the satisfiability problem is polynomial. Certainly, further studies of the realizability problem would be worthwhile.

Another obvious research direction is to completely classify other spatial algebras, such as RCC-8. RCC-8 contains $2^{256} \approx 10^{77}$ relations so the question whether this is feasible or not remains to be answered.

## 6. Conclusions

We have studied computational properties of RCC-5. All of the $2^{32}$ possible subclasses are classified with respect to whether their corresponding satisfiability problem is tractable or not. The classification reveals that there are four maximal tractable subclasses of the algebra.